\documentclass[runningheads,a4paper]{llncs}
\usepackage{latexsym}
\usepackage{lipsum}
\usepackage[section]{placeins}
\usepackage{floatrow}
\usepackage{booktabs}
 \usepackage{multirow}
\usepackage[section]{placeins}	

\usepackage{float}
\floatstyle{plaintop}
\restylefloat{table}
\usepackage{framed}

\usepackage{amsmath}
\usepackage{amssymb}

\usepackage{algorithm}
\usepackage[noend]{algpseudocode}

\usepackage{times}
\usepackage{graphicx}

\usepackage{url}
\urldef{\mailsa}\path|{nsanchan1, ahmet.aker, k.bontcheva}@sheffield.ac.uk|    
\newcommand{\keywords}[1]{\par\addvspace\baselineskip
\noindent\keywordname\enspace\ignorespaces#1}

\makeatletter
\def\BState{\State\hskip-\ALG@thistlm}
\makeatother

\begin{document}

\mainmatter  

\title{Gold Standard Online Debates Summaries and First Experiments Towards Automatic Summarization of Online Debate Data}

\titlerunning{Gold Standard Online Debates Summaries}

%
%
\author{Nattapong Sanchan \and Ahmet Aker \and Kalina Bontcheva}
\authorrunning{Gold Standard Online Debates Summaries}

\institute{Natural Language Processing Group, Department of Computer Science,\\
The University of Sheffield, 211 Portobello, Sheffield, United Kingdom\\
\mailsa\\
\url{https://www.sheffield.ac.uk/dcs}}

%
%

\toctitle{Lecture Notes in Computer Science}
\tocauthor{Authors' Instructions}
\maketitle

\pagenumbering{gobble} 

%
%
%
\begin{abstract} 

Usage of online textual media is steadily increasing. Daily, more and more news stories, blog posts and scientific articles are added to the online volumes. These are all freely accessible and have been employed extensively in multiple research areas, e.g. automatic text summarization, information retrieval, information extraction, etc. Meanwhile, online debate forums have recently become popular, but have remained largely unexplored. For this reason, there are no sufficient resources of annotated debate data available for conducting research in this genre. In this paper, we collected and annotated debate data for an automatic summarization task. Similar to extractive gold standard summary generation our data contains sentences worthy to include into a summary. Five human annotators performed this task. Inter-annotator agreement, based on semantic similarity, is 36\% for Cohen's kappa and 48\% for Krippendorff's alpha. Moreover, we also implement an extractive summarization system for online debates and discuss prominent features for the task of summarizing online debate data automatically.

\end{abstract}

%
%
%
\section{Introduction}
With the exponential growth of Internet usage, online users massively publish textual content on online media. For instance, a micro-blogging website, Twitter, allows users to post their content in 140-characters length. A popular social media like Facebook allows users to interact and share content in their communities, as known as ``Friends''. An electronic commercial website, Amazon, allows users to ask questions on their interested items and give reviews on their purchased products. While these textual data have been broadly studied in various research areas (e.g. automatic text summarization, information retrieval, information extraction, etc.), online debate domain, which recently becomes popular among Internet users, has not yet largely explored. For this reason, there are no sufficient resources of annotated debate data available for conducting research in this genre. This motivates us to explore online debate data. 

In this paper, we collected and annotated debate data for an automatic summarization task. There are 11 debate topics collected. Each topic consists of different number of debate comments. In total, there are 341 debate comments collected, accounting for 2518 sentences. In order to annotate online debate data, we developed a web-based system which simply runs on web browsers. We designed the user interface for non-technical users. When participants logged into the system, a debate topic and a comment which is split to a list of consecutive sentences were shown at a time. The annotators were asked to select salient sentences from each comment which summarize it. The number of salient sentences chosen from each comment is controlled by a compression rate of 20\% which is automatically calculated by the web-based system. For instance, Table \ref{table_annotation} shows a debate comment to be annotated by an annotator. Based on the compression rate of 20\%, the annotator needs to choose 1 sentence that summarizes the comment. This compression rate was also used in \cite{Neto2002ATS} and \cite{Morris199217}. In total, we obtained 5 sets of annotated debate data. Each set of data consists of 341 comments with total 519 annotated salient sentences.

Inter-annotator agreement in terms of Cohen's Kappa and Krippendorff's alpha are 0.28 and 0.27 respectively. For social media data such low agreements have been also reported by related work. For instance, \cite{Mitrat} reports Kappa scores between 0.20 and 0.50 for human constructed newswire summaries. \cite{Liu:2008:CRH:1557690.1557747} reports again Kappa scores between 0.10 and 0.35 for the conversation transcripts. Our agreement scores are based on strict conditions where agreement is achieved when annotators have selected exact the same sentences. However, such condition does not consider syntactically different sentences bearing the same semantic meaning. Thus we also experimented with a more relaxed version that is based on semantic similarity between sentences. We regard two sentences as identical when their semantic similarity is above a threshold. Our results revealed that after applying such an approach the averaged Cohen's Kappa and Krippendorff's alpha increase to 35.71\% and 48.15\% respectively.

Finally we report our results of automatic debate data summarization. We implemented an extractive text summarization system that extracts salience sentences from user comments. Among the features the most contributing ones are sentence position, debate titles, and cosine similarity of the debate title words and sentences.

The paper is structured as follows. First we describe the nature of our online debate data. In Section \ref{data_annotation} we discuss the procedures of data annotation and discuss our experiments with semantic similarity applied on inter-annotator agreement computation. In Section \ref{experiment_salient}, we present our first results on automatically performing debate data summarization. We conclude in Section \ref{conclusion}.

\begin{table}[ht]
\begin{flushleft}
\begin{framed}
\noindent\textbf{Task 02: Is global warming fictitious?}\\
\emph{$[1]$} I do not think global warming is fictitious.\\
\emph{$[2]$} I understand a lot of people do not trust every source and they need solid proof.\\
\emph{$[3]$} However, if you look around us the proof is everywhere.\\
\emph{$[4]$} It began when the seasons started getting harsh and the water levels were rising.\\
\emph{$[5]$} I do not need to go and see the ice caps melting to know the water levels are rising and the weather is changing.\\
\emph{$[6]$} I believe global warming is true, and we should try and preserve as much of the Earth as possible.
\end{framed}
\end{flushleft}
\caption{Examples of the debate data to be annotated.}\label{table_annotation} 
\end{table}

%
%
%
\begin{table}[ht]
\begin{flushleft}
\begin{framed}
\textbf{Example 1: Propositions from the proponents}  
\\ - Global warming is real.
\\ - Global warming is an undisputed scientific fact.     
\\ - Global warming is most definitely not a figment of anyone's imagination, because the proof is all around us.
\\ - I believe that global warming is not fictitious, based on the observational and comparative evidence that is currently presented to us.
\vskip 0.2in
\textbf{Example 2: Propositions from the opponents}  
\\ - Global warming is bull crap.
\\ - Global Warming isn't a problem at all.
\\ - Just a way for the government to tax people on more things by saying their trying to save energy.
\\ - Yes, global warming is a myth, because they have not really proven the science behind it. 
\\  
\end{framed}
\end{flushleft}
\caption{Examples of Paraphrased Arguments.}\label{table_pargument} 
\end{table}
\FloatBarrier

\section{Online Debate Data and Their Nature} \label{nature_debate}

The nature of online debate is different from other domains. It gives opportunities to users to discuss ideological debates in which users can choose a stance of a debate, express their opinions to support their stance, and oppose other stances. To conduct our experiments we collected debate data from the Debate discussion forum.\footnote{http://www.debate.org} The data are related to an issue of the existence of global warming. In the data, there are two main opposing sides of the arguments. A side of proponents believes in the existence of global warming and the other side, the opponents, says that global warming is not true. When the proponents and the opponents express their sentiments, opinions, and evidences to support their propositions, the arguments between them arise. Moreover, when the arguments are referred across the conversation in the forum, they are frequently paraphrased. Table \ref{table_pargument} illustrates examples of the arguments being paraphrased. Sentences expressing related meaning are written in different context.

%
%
%
\section{Annotation Procedures} \label{data_annotation}
In this paper, we collected and annotated debate data for an automatic summarization task. There are 11 debate topics collected. Each topic consists of a different number of debate comments as shown in Table \ref{table_stats_corpus}. The annotation was guided through a web-based application. The application was designed for non-technical users. When participants logged in to the system, a debate topic and a comment which is split to a list of sentences were shown at a time. The annotators were given a guideline to read and select salient sentences that summarize the comments. From each comment we allowed the participants to select only 20\% of the comment sentences. These 20\% of the sentences are treated as the summary of the shown comment. In the annotation task, all comments in the 11 debate topics were annotated. We recruited 22 participants: 10 males and 12 participants to annotate salient sentences. The participants' backgrounds were those who are fluent in English and aged above 18 years old. We aimed to have 5 annotations sets for each debate topic. Due to a limited number of annotators and a long list of comments to be annotated in each debate topic, 11 participants were asked to complete more than one debate topic, but were not allowed to annotate the same debate topics in which they had done before. In total, 55 annotation sets were derived: 11 debate topics and each with 5 annotation sets. Each annotation set consists of 341 comments with total 519 annotated salient sentences.\footnote{This dataset can be downloaded at https://goo.gl/3aicDN.}

\begin{table}[ht]
\centering
\begin{tabular}{@{}clccc@{}}
\toprule
\textbf{Topic ID} & \multicolumn{1}{c}{\textbf{Debate Topics}} & \textbf{Comments} & \textbf{Sentences} & \textbf{Words} \\ \midrule
01 & Is global warming a myth? & 18 & 128 & 2701 \\
02 & Is global warming fictitious? & 28 & 173 & 3346 \\
03 & Is the global climate change man made? & 10 & 47 & 1112 \\
04 & Is global climate change man-made? & 103 & 665 & 12054 \\
05 & Is climate change man-made? & 9 & 46 & 773 \\
06 & Do you believe in global warming? & 21 & 224 & 3538 \\
07 & Does global warming exist? & 68 & 534 & 9178 \\
08 & \begin{tabular}[c]{@{}l@{}}Can someone prove that climate \\ change is real (yes) or fake (no)?\end{tabular} & 8 & 49 & 1127 \\
09 & Is global warming real? & 51 & 434 & 6749 \\
10 & Is global warming true? & 5 & 26 & 375 \\
11 & \begin{tabular}[c]{@{}l@{}}Is global warming real (yes) or just a bunch \\ of scientist going to extremes (no)?\end{tabular} & 20 & 192 & 2988 \\\midrule
\textbf{} & \multicolumn{1}{r}{\textbf{Average}} & \textbf{31} & \textbf{229} & \textbf{3995} \\
\textbf{} & \multicolumn{1}{r}{\textbf{Total}} & \textbf{341} & \textbf{2518} & \textbf{43941} \\ \bottomrule
\end{tabular}
\caption{Statistical information of the online debate corpus.}
\label{table_stats_corpus}
\end{table}

\subsection{Inter-Annotator Agreement}

In order to compute inter-annotator agreement between the annotators we calculated the averaged Cohen's Kappa and Krippendorff's alpha with a distant metric, Measuring Agreement on Set-valued Items metric (MASI). The scores of averaged Cohen's Kappa and Krippendorff's alpha are 0.28 and 0.27 respectively. According to the scale of \cite{krippendorff-2004}, our alpha did neither accomplish the reliability scale of 0.80, nor the marginal scales between 0.667 and 0.80. Likewise, our Cohen's Kappa only achieved the agreement level of \emph{fair agreement}, as defined by \cite{Landis77}. However, such low agreement scores are also reported by others who aimed creating gold standard summaries from news texts or conversational data \cite{Mitrat}  \cite{Liu:2008:CRH:1557690.1557747} .

Our analysis shows that the low agreement is caused by different preferences of annotators in the selection of salient sentences. As shown in Table \ref{table_pargument} the sentences are syntactically different but bear the same semantic meaning. In a summarization task with a compression threshold, such situation causes the annotators to select one of the sentences but not all. Depending on each annotator's preference the selection leads to different set of salient sentences. To address this we relaxed the agreement computation by treating sentences equal when they are semantically similar. We outline details in the following section.

\subsection{Relaxed Inter-Annotator Agreement}

When an annotator selects a sentence, other annotators might select other sentences expressing similar meaning. In this experiment, we aim to detect sentences that are semantically similar by applying Doc2Vec from the Gensim package \cite{rehurek_lrec}. Doc2Vec model simultaneously learns the representation of words in sentences and the labels of the sentences. The labels are numbers or chunks of text which are used to uniquely identify each sentence. We used the debate data and a richer collections of sentences related to climate change to train the Doc2Vec model. In total, there are 10,920 sentences used as the training set. 

To measure how two sentences are semantically referring to the same content, we used a function provided in the package to calculate cosine similarity scores among sentences. A cosine similarity score of 1 means that the two sentences are semantically equal and 0 is when it is opposite the case. In the experiment, we manually investigated pairs of sentences at different threshold values and found that the approach is stable at the threshold level above 0.44. The example below shows a pair of sentences obtained at 0.44 level. \\

\indent \textbf{S1: }\emph{Humans are emitting carbon from our cars, planes and factories, which is a heat trapping particle.}\\
\indent \textbf{S2: }\emph{So there is no doubt that carbon is a heat trapping particle, there is no doubt that our actions are emitting carbon into the air, and there is no doubt that the amount of carbon is increasing.}\\

In the pair, the two sentences mention the same topic (i.e. \emph{carbon emission}) and express the idea in the same context. We used the threshold 0.44 to re-compute the agreement scores. By applying the semantic approach, the inter-annotator agreement scores of Cohen's Kappa and Krippendorff's alpha increase from 0.28 to 35.71\% and from 0.27 to 48.15\% respectively. The inter-annotator agreement results are illustrated in Table \ref{iaa}. Note that, in the calculation of the agreement, we incremented the threshold by 0.02. Only particular thresholds are shown in the table due to the limited space.

\begin{table}[ht]
\centering
\begin{tabular}{@{}lcccc@{}}
\toprule
\multicolumn{1}{c}{\textbf{Trial}} & \textbf{\begin{tabular}[c]{@{}c@{}}Threshold\\ ($\ge$)\end{tabular}} & \textbf{$\kappa$} & &\textbf{$\alpha$} \\ \midrule
\multicolumn{1}{c}{Before} &  & 0.28  && 0.27 \\\midrule
\multicolumn{1}{c}{After} & 0.00 & 0.81  && 0.83 \\ 
 & 0.10 & 0.62 & & 0.65 \\
 & 0.20 & 0.46 & & 0.50 \\
 & 0.30 & 0.40 & & 0.43 \\
 & 0.40 & 0.39 & & 0.41 \\
 & 0.42 & 0.38 & & 0.41 \\
 & \textbf{0.44} & \textbf{0.38} & & \textbf{0.40} \\
 & 0.46 & 0.38 & & 0.40 \\
 & 0.48 & 0.38 & & 0.40 \\
 & 0.50 & 0.38 & & 0.40 \\
 & 0.60 & 0.38 & & 0.40 \\
 & 0.70 & 0.38 & & 0.40 \\
 & 0.80 & 0.38 & & 0.40 \\
 & 0.90 & 0.38 & & 0.40 \\
 & 1.00 & 0.38 & & 0.40 \\\bottomrule
\end{tabular}
\caption{Inter-Annotator Agreement before and after applying the semantic similarity approach.}
\label{iaa}
\end{table}

%
%
%

\section{Automatic Salient Sentence Selection} \label{experiment_salient} 
\subsection{Support Vector Regression Model}

 In this experiment, we work on extractive summarization problem and aim to select sentences that are deemed important or that summarize the information mentioned in debate comments. Additionally, we aim to investigate the keys features which play the important roles in the summarization of the debate data. We view this salient sentence selection as a regression task. A regression score for each sentence is ranged between 1 to 5. It is derived by the number annotators selected that sentence divided by the number of all annotators. In this experiment, a popular machine learning package which is available in Python, called Scikit-learn \cite{scikitLearn} is used to build a support vector regression model. We defined 8 different features and the support vector regression model combines the features for scoring sentences in each debate comment. From each comment, sentences with the highest regression scores are considered the most salient ones. 

\subsection{Feature Definition}
\begin{enumerate}
    \item \textbf{Sentence Position (SP).}
Sentence position correlates with the important information in text \cite{Baxendale,EdmundsonRatingSummary,Goldstein}. In general, humans are likely to mention the first topic in the earlier sentence and they express more information about it in the later sentences. We prove this claim by conducting a small experiment to investigate which sentence positions frequently contain salient sentences. From our annotated data, the majority votes of the sentences are significantly at the first three positions (approximately 60\%), shaping the assumption that the first three sentences are considered as containing salient pieces of information. Equation \ref{eq_sentence_position} shows the calculation of the score obtained by the sentence position feature.  \\
    \begin{equation}	 \label{eq_sentence_position}
      SP=\left\{
      \begin{array}{@{}ll@{}}
        \frac{1}{sentence \; position}, & \text{if}\ position <4 \\
        0, & \text{otherwise}
      \end{array}\right.
     \end{equation}

\item \textbf{Debate Titles (TT).}
In writing, a writer tends to repeat the title words in a document. For this reason, a sentence containing title words is likely to contain important information. We collected 11 debate titles as shown in Table \ref{table_stats_corpus}. In our experiment, a sentence is considered as important when it contains mutual words as in debate titles. Equation \ref{eq_titleword} shows the calculation of the score by this feature. \\
    \begin{equation} \label{eq_titleword}
    TT  = \frac{\; number \; of \; title \; words \; in \; sentence}{number \; of \; words \;in \;debate \;titles}
    \end{equation}

    \item \textbf{Sentence Length (SL).} 
Sentence length also indicates the importance of sentence based on the assumption that either very short or very long sentences are unlikely to be included in the summary. Equation \ref{eq_sentencelength} is used in the process of extracting salient sentences from debate comments. \\
    \begin{equation} \label{eq_sentencelength}
    SL  = \frac{\; number \; of \; words \; in \; a \; sentence}{number \; of \; words\; in \; the \; longest \; sentence}
    \end{equation}

    \item \textbf{Conjunctive Adverbs (CJ).}
One possible feature that helps identify salient sentence is to determine conjunctive adverbs in sentences. Conjunctive adverbs were proved that they support cohesive structure of writing. For instance, ``the conjunctive adverb \emph{moreover} has been used mostly in the essays which lead to a conclusion that it is one of the best accepted linkers in the academic writing process." \cite{januliene2015use}. The NLTK POS Tagger\footnote{http://www.nltk.org/api/nltk.tag.html} was used to determine conjunctive adverbs in our data. \\

    \item \textbf{Cosine Similarity.}
Cosine similarity has been used extensively in Information Retrieval, especially in the vector space model. Documents will be ranked according to the similarity of the given query. Equation \ref{cosinesim} illustrates the equation of cosine similarity, where: \emph{q} and \emph{d} are n-dimensional vectors \cite{Manning:1999:FSN:311445}. Cosine similarity is one of our features that is used to find similarity between two textual units. The following features are computed by applying cosine similarity. 

    \begin{equation} \label{cosinesim}		
    cos(q,d) = \frac{\sum\limits_{i=1}^n q_{i} d_{i}}{\sqrt{\sum\limits_{i=1}^n q^2_{i}}\sqrt{\sum\limits_{i=1}^n d^2_{i}}} 
    \end{equation}

        \begin{enumerate}
            \item \textbf{Cosine similarity of debate title words and sentences (COS\_TTS).} For each sentence in debate comments we compute its cosine similarity score with the title words. This is based on the assumption that a sentence containing title words is deemed as important. \\
            
            \item \textbf{Cosine similarity of climate change terms and sentences (COS\_CCTS)}. The climate change terms were collected from news media about climate change. We calculate cosine similarity between the terms and sentences. In total, there are 300 most frequent terms relating to location, person, organization, and chemical compounds.\\
            
             \item \textbf{Cosine similarity of topic signatures and sentences (COS\_TPS).} Topic signatures play an important role in automatic text summarization and information retrieval. It helps identify the presence of complex concepts or the importance in text. In a process of determining topic signatures, words appearing occasionally in the input text but rarely in other text are considered as topic signatures. They are determined by an automatic predefined threshold which indicates descriptive information. Topic signatures are generated by comparing with pre-classified text on the same topic using a concept of likelihood ratio \cite{nenkova-mckeown-2011,Lin:2000:AAT:990820.990892}, $\lambda$ presented by \cite{Dunning1993}. It is a statistical approach which calculates a likelihood of a word. For each word in the input, the likelihood of word occurrence is calculated in pre-classified text collection. Another likelihood values of the same word is calculated and compared in another out-of-topic collection. The word, on the topic-text collection that has higher likelihood value than the out-of-topic collection, is regarded as topic signature of a topic. Otherwise the word is ignored. \\        
        \end{enumerate}
    \item \textbf{Semantic Similarity of Sentence and Debate Titles (COS\_STT).} Since the aforementioned features do not semantically capture the meaning of context, we create this feature for such purpose. We compare each sentence to the list of debate titles based on the assumption that forum users are likely to repeat debate titles in their comments. Thus, we compare each sentence to the titles and then calculate the semantic similarity score by using Doc2Vec \cite{rehurek_lrec}.   
\end{enumerate}

\begin{table}[ht]
\centering
\scalebox{0.85}{
\begin{tabular}{|c|c|c|c|c|c|c|c|c|c|}
\hline

\textbf{ROUGE-N} & \textbf{CB} & \textbf{CJ} & \textbf{COS\_CCT} & \textbf{COS\_TTS} & \textbf{COS\_TPS} & \textbf{SL} & \textbf{SP} & \textbf{COS\_STT} & \textbf{TT} \\ \hline
\textbf{R-1} & 0.4773 & 0.4988 & 0.3389 & 0.5630 & 0.3907 & 0.4307 & \textbf{0.6124} & 0.4304 & 0.5407 \\ \hline
\textbf{R-2} & 0.3981 & 0.4346 & 0.2558 & 0.5076 & 0.2986 & 0.3550 & \textbf{0.5375} & 0.3561 & 0.4693 \\ \hline
\textbf{R-SU4} & 0.3783 & 0.4147 & 0.2340 & 0.4780 & 0.2699 & 0.3335 & \textbf{0.4871} & 0.3340 & 0.4303 \\ \hline
\end{tabular}}
\caption{ROUGE scores after applying Doc2Vec to the salient sentence selection.}\label{table_rouge_scores}
\end{table}

\begin{table}[ht]
\centering
\scalebox{0.85}{
\begin{tabular}{lcccccc}
\hline
\multicolumn{1}{|c|}{\multirow{2}{*}{\textbf{Comparison Pairs}}} & \multicolumn{2}{c|}{\textbf{ROUGE-1}} & \multicolumn{2}{c|}{\textbf{ROUGE-2}} & \multicolumn{2}{c|}{\textbf{ROUGE SU4}} \\ \cline{2-7} 
\multicolumn{1}{|c|}{} & \multicolumn{1}{c|}{\textbf{Z}} & \multicolumn{1}{c|}{\textbf{\begin{tabular}[c]{@{}c@{}}Asymp. Sig.\\ (2-tailed)\end{tabular}}} & \multicolumn{1}{c|}{\textbf{Z}} & \multicolumn{1}{c|}{\textbf{\begin{tabular}[c]{@{}c@{}}Asymp. Sig.\\ (2-tailed)\end{tabular}}} & \multicolumn{1}{c|}{\textbf{Z}} & \multicolumn{1}{c|}{\textbf{\begin{tabular}[c]{@{}c@{}}Asymp. Sig.\\ (2-tailed)\end{tabular}}} \\ \hline
\multicolumn{1}{|l|}{SP VS CB} & \multicolumn{1}{c|}{$-4.246^b$} & \multicolumn{1}{c|}{0*} & \multicolumn{1}{c|}{$-3.962^b$} & \multicolumn{1}{c|}{0*} & \multicolumn{1}{c|}{$-3.044^b$} & \multicolumn{1}{c|}{0.002} \\ \hline
\multicolumn{1}{|l|}{SP VS CJ} & \multicolumn{1}{c|}{$-3.570^b$} & \multicolumn{1}{c|}{0*} & \multicolumn{1}{c|}{$-3.090^b$} & \multicolumn{1}{c|}{0.002} & \multicolumn{1}{c|}{$-2.192^b$} & \multicolumn{1}{c|}{0.028} \\ \hline
\multicolumn{1}{|l|}{SP VS COS\_CCTS} & \multicolumn{1}{c|}{$-6.792^b$} & \multicolumn{1}{c|}{0*} & \multicolumn{1}{c|}{$-6.511^b$} & \multicolumn{1}{c|}{0*} & \multicolumn{1}{c|}{$-6.117^b$} & \multicolumn{1}{c|}{0*} \\ \hline
\multicolumn{1}{|l|}{SP VS COS\_TTS} & \multicolumn{1}{c|}{$-1.307^b$} & \multicolumn{1}{c|}{0.191} & \multicolumn{1}{c|}{$-.789^b$} & \multicolumn{1}{c|}{0.43} & \multicolumn{1}{c|}{$-.215^b$} & \multicolumn{1}{c|}{0.83} \\ \hline
\multicolumn{1}{|l|}{SP VS COS\_TPS} & \multicolumn{1}{c|}{$-6.728^b$} & \multicolumn{1}{c|}{0*} & \multicolumn{1}{c|}{$-6.663^b$} & \multicolumn{1}{c|}{0*} & \multicolumn{1}{c|}{$-6.384^b$} & \multicolumn{1}{c|}{0*} \\ \hline
\multicolumn{1}{|l|}{SP VS SL} & \multicolumn{1}{c|}{$-4.958^b$} & \multicolumn{1}{c|}{0*} & \multicolumn{1}{c|}{$-4.789^b$} & \multicolumn{1}{c|}{0*} & \multicolumn{1}{c|}{$-4.110^b$} & \multicolumn{1}{c|}{0*} \\ \hline
\multicolumn{1}{|l|}{SP VS COS\_STT} & \multicolumn{1}{c|}{$-4.546^c$} & \multicolumn{1}{c|}{0*} & \multicolumn{1}{c|}{$-4.322^c$} & \multicolumn{1}{c|}{0*} & \multicolumn{1}{c|}{$-3.671^c$} & \multicolumn{1}{c|}{0*} \\ \hline
\multicolumn{1}{|l|}{SP VS TT} & \multicolumn{1}{c|}{$-3.360^c$} & \multicolumn{1}{c|}{0.001*} & \multicolumn{1}{c|}{$-2.744^c$} & \multicolumn{1}{c|}{0.006} & \multicolumn{1}{c|}{$-2.641^c$} & \multicolumn{1}{c|}{0.008} \\ \hline
\multicolumn{7}{l}{a) Wilcoxon Signed Ranks Test.} \\
\multicolumn{7}{l}{b) Based on negative ranks.} \\
\multicolumn{7}{l}{c) Based on positive ranks.}
\end{tabular}}
\caption{The statistical information of comparing sentence position and other features after applying Doc2Vec.}
\label{table_sig_position}
\end{table}

\subsection{Results} \label{results}
In order to evaluate the system summaries against the reference summaries, we apply ROUGE-N evaluation metrics. We report ROUGE-1 (unigram), ROUGE-2 (bi-grams) and ROUGE-SU4 (skip-bigram with maximum gap length of 4). The ROUGE scores as shown in Table \ref{table_rouge_scores} indicate that sentence position feature outperforms other features. The least performing feature is the cosine similarity of climate change terms and sentences feature.

To measure the statistical significance of the ROUGE scores generated by the features, we calculated a pairwise Wilcoxon signed-rank test with Bonferroni correction. We report the significance p = .0013 level of significance after the correct is applied. Our results indicate that there is statistically significance among the features. Table \ref{table_sig_position} illustrates the statistical information of comparing sentence position and other features. The star indicates that there is a statistical significance difference between each comparison pair.

%
%
%
\section{Conclusion} \label{conclusion}
In this paper we worked on an annotation task for a new annotated dataset, online debate data. We have manually collected reference summaries for comments given to global warming topics. The data consists of 341 comments with total 519 annotated salient sentences. We have performed five annotation sets on this data so that in total we have 5 X 519 annotated salient sentences. We also implemented an extractive text summarization system on this debate data. Our results revealed that the key feature that plays the most important role in the selection salient sentences is sentence position. Other useful features are debate title words feature, and cosine similarity of debate title words and sentences feature. 

In future work, we aim to investigate further features for the summarization purposes. We also plan to integrate stance information so that summaries with pro-contra sides can be generated.

\section*{Acknowledgments}
This work was partially supported by the UK EPSRC Grant No. EP/I004327/1, the European Union under Grant Agreements No. 611233 PHEME, and the authors would like to thank Bankok University of their support.

\end{document}